\def\year{2019}\relax
\documentclass[letterpaper]{article} 
\usepackage{aaai19}  
\usepackage{times}  
\usepackage{helvet}  
\usepackage{courier}  
\usepackage{url}  
\usepackage{graphicx}  

\usepackage{amsthm}
\usepackage{amsmath}
\usepackage{multirow}
\usepackage{booktabs} 

\begin{document}
%
\title{Multi-Level Deep Cascade Trees for Conversion Rate Prediction in Recommendation System}

\author{
Hong Wen{$^{1}$~$^{*}$},
Jing Zhang{$^{2,3}$~\thanks{The co-first authors contributed equally.}},
Quan Lin{$^{1}$},
Keping Yang{$^{1}$~\thanks{Corresponding author.}},
Pipei Huang{$^{1}$} \\
{$^{1}$} Alibaba Group \\
{$^{2}$} Hangzhou Dianzi University \\
{$^{3}$} University of Technology Sydney \\
\{qinggan.wh, pipei.hpp\}@alibaba-inc.com \\
\{tieyi.lq, shaoyao\}@taobao.com, jing.zhang@uts.edu.au
}

\maketitle

\begin{abstract}
Developing effective and efficient recommendation methods is very challenging for modern e-commerce platforms. Generally speaking, two essential modules named ``Click-Through Rate Prediction'' (\textit{CTR}) and ``Conversion Rate Prediction'' (\textit{CVR}) are included, where \textit{CVR} module is a crucial factor that affects the final purchasing volume directly. However, it is indeed very challenging due to its sparseness nature. In this paper, we tackle this problem by proposing multi-Level Deep Cascade Trees (\textit{ldcTree}), which is a novel decision tree ensemble approach. It leverages deep cascade structures by stacking Gradient Boosting Decision Trees (\textit{GBDT}) to effectively learn feature representation. In addition, we propose to utilize the cross-entropy in each tree of the preceding \textit{GBDT} as the input feature representation for next level \textit{GBDT}, which has a clear explanation, i.e., a traversal from root to leaf nodes in the next level \textit{GBDT} corresponds to the combination of certain traversals in the preceding \textit{GBDT}. The deep cascade structure and the combination rule enable the proposed \textit{ldcTree} to have a stronger distributed feature representation ability. Moreover, inspired by ensemble learning, we propose an Ensemble \textit{ldcTree} (\textit{E-ldcTree}) to encourage the model's diversity and enhance the representation ability further. Finally, we propose an improved Feature learning method based on \textit{EldcTree} (\textit{F-EldcTree}) for taking adequate use of weak and strong correlation features identified by pre-trained \textit{GBDT} models. Experimental results on off-line data set and online deployment demonstrate the effectiveness of the proposed methods.
\end{abstract}

\section{Introduction}

With the explosive growth of information available online, Recommender System (\textit{RS}), as a useful information filtering tool, is used for estimating users' preferences on items they have not seen and guides them to discover products or services they might be interested in from massive possible options. In general, recommender systems are classified into the following three categories based on the forms of recommendations \cite{balabanovic1997fab}: Collaborative recommendations, Content-based recommendations and Hybrid recommendations. Collaborative recommendations make users recommended items that people with similar tastes preferred in the past. Content-based recommendations make users recommended items similar to the ones the user preferred in the past. Hybrid recommendations integrates two or more types of recommendation strategies, which helps to avoid certain limitations of single strategy \cite{adomavicius2005toward}.

\begin{figure}
\centering
\includegraphics[width=0.8\linewidth]{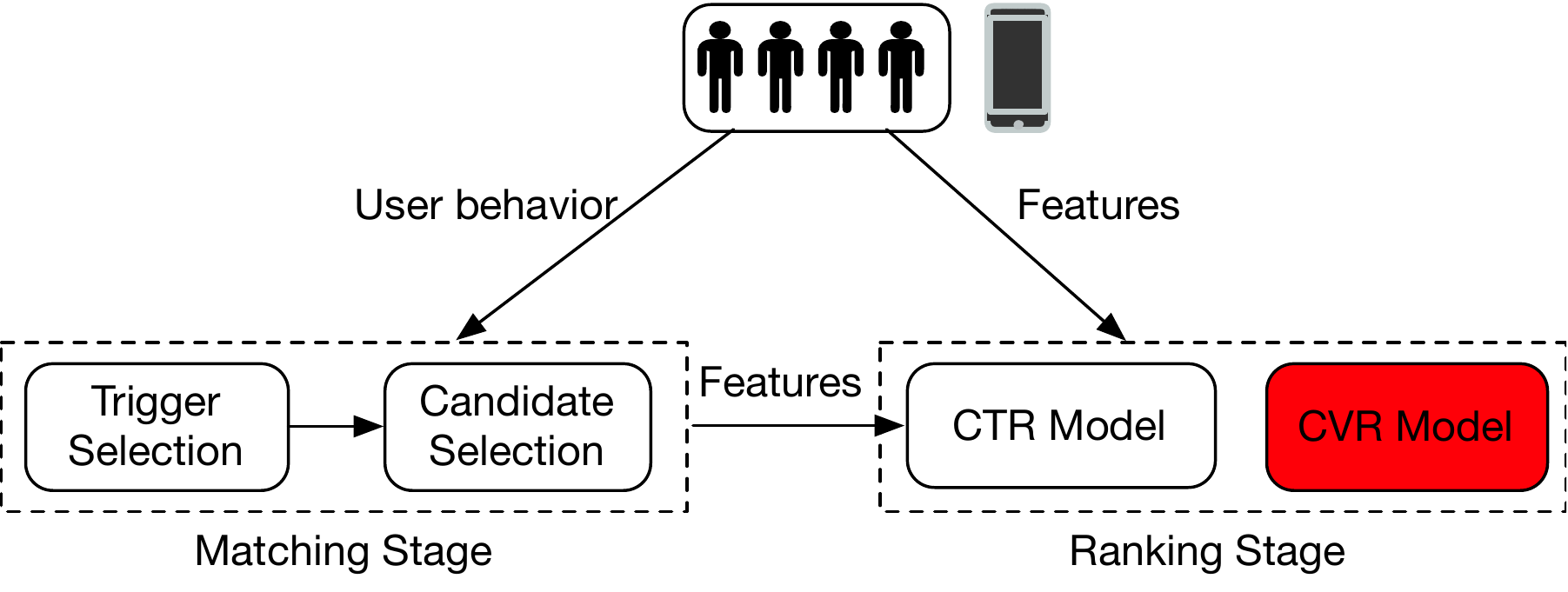}
\caption{The framework for online recommendation in our E-commerce platform.}
\label{fig:flow}
\end{figure}

Recommender System increasingly plays an essential role in industry area, which promotes services for many applications \cite{gomez2016netflix,davidson2010youtube}. In addition, in order to help customers find exactly what they need, recommendation techniques have been studied and deployed extensively on E-commerce platforms, which provides good user experience and promotes incredible increment in revenue. Usually, the deployed framework for our online E-commerce platforms is illustrated in Fig. \ref{fig:flow}. Specifically, when a user visits it through a terminal, such as smart phones, the system firstly analyzes his/her long and short term behaviors and then his/her interested items, called \textit{Triggers}, are selected. Then, massive items closely related with \textit{Triggers} are generated. Further, top K(e.g., 500) of them (based on the \textit{``matching score''}), along with extra information (e.g., user, item, user-item cross features), are delivered to the next \textit{Ranking} stage where it mainly contains two core modules, namely \textit{CTR} and \textit{CVR}. Finally, the recommendation results are generated and displayed to the user. In this paper, we mainly focused on the \textit{CVR} module.

In the past few decades, deep learning has been witnessed the tremendous successes in many application areas \cite{zhang2017deep,liu2017survey}, such as image classification \cite{krizhevsky2012imagenet,simonyan2014very} , speech recognition \cite{deng2010binary,deng2013recent} and object detection \cite{girshick2015fast,ren2015faster}. Meanwhile, recent studies also demonstrate its efficiency and effectiveness in coping with recommendation tasks\cite{elkahky2015multi,wang2015collaborative,chen2017attentive,he2017neural,huang2015neural,yang2017bridging,guo2017deepfm}. Though deep learning has been partially overcoming obstacles of conventional models and gaining momentum due to its state-of-the-art performances, it has apparent deficiencies, such as a huge amount of data and powerful computational facilities required for training, more importantly many hyper-parameters to be tuned. Recently, \textit{gcForest} \cite{zhou2017deep}, an alternative to \textit{DNN}, is proposed, which generates a deep forest ensemble, with a cascade structure to do representation learning. In addition, it achieves highly competitive performance compared with \textit{DNN} for various domains' tasks while having fewer hyper-parameters.

In this paper, partially inspired by \textit{gcForest} \cite{zhou2017deep}, we firstly propose a multi-Level Deep Cascade Trees model (short as \textit{ldcTree}) to cope with the essential task \textit{CVR prediction} in the ``also view'' module. \textit{ldcTree} is another alternative to \textit{DNN} and encourages to do representation learning by a level-by-level cascade structure. Specifically, it takes a multi-dimensional representational feature vector from preceding level and outputs its processing results to the next level by employing Gradient Boosting Decision Trees (\textit{GBDT} ) models \cite{friedman2001greedy}, where a new tree is created to model the residual of previous trees during each iteration, and a traversal from root node to a leaf node represents a combination rule of certain input features. One step further, we propose to utilize the cross-entropy value of each leaf node on the trees in the preceding level \textit{GBDT} to construct the feature representation for the next level \textit{GBDT}, which results a clear explanation of the spliting node in the next level \textit{GBDT}, i.e., a traversal from root to a leaf node in next level \textit{GBDT} indicates a combination rule of certain paths on the trees from preceding level \textit{GBDT}. Then, inspired by the idea of ensemble learning, we proposed Ensemble \textit{ldcTree} (short as \textit{E-ldcTree}), which encourages the model's diversity and enhances the representation ability.

Furthermore, it is noteworthy that a small number of raw features contributes the majority of explanatory power while the remaining features have only a marginal contribution in \textit{GBDT} models \cite{he2014practical}, which results in the importance of certain raw features can't be demonstrated. Therefore, we further proposed an improved Feature learning method based on the above \textit{EldcTree}, named \textit{F-EldcTree}, which takes more adequately use of weak and strong correlation features identified by pre-trained \textit{GBDT} model at corresponding levels. The key contributions of this paper are:

\begin{itemize}
\item{We propose a novel model \textit{ldcTree} and its extension $EldcTree$, which are decision tree ensemble methods by exploiting the deep cascade structures and using a cross-entropy based feature representation. It exhibits a strong feature representation ability and has a clear explanation.}
\item{To the extent of our knowledge, our proposed \textit{F-EldcTree} is the first recommendation work which adequately takes into account weak and strong correlation features identified by pre-trained \textit{GBDT} model at corresponding levels, and contributes more excellent ability for representation learning.}
\item{We have successfully deployed the proposed methods to the recommendation module in our E-commerce platform, and carry out online experiments with more than 100M users and items, furthermore achieves 12 percent \textit{CVR} improvement compared with the baseline model.}
\end{itemize}

The rest of this paper is organized as follows. Section 2 briefly reviews existing related work. Section 3 describes the proposed approach in detail, followed by presenting experimental results on both off-line evaluation and online applications in Section 4. We conclude the paper in Section 5.

\section{Related work}

\subsection{Conversion Rate Prediction}

Conversions are very rare events and only a very small portion of the users will eventually convert after clicking or being shown, resulting in extremely challenging for building thus models \cite{mahdian2007pay,chapelle2015simple,rosales2012post,oentaryo2014predicting}. In addition, it can be broadly divided to two categories of Post View Conversion (\textit{PVC}) and Post Click Conversion (\textit{PCC}), which means conversion after viewing an item without having clicked it itself and conversion after having clicked, respectively. In the context of this paper, conversion refers to the purchase event that occurs after a user clicking an item, i.e. post click conversions (\textit{PCC}) \cite{rosales2012post}.

\subsection{Tree based Feature Representation}

\textit{GBDT} follows the Gradient Boosting Machine (\textit{GBM}) \cite{friedman2001greedy}, which produces competitive, highly robust, interpretable procedures for both regression and classification, especially appropriate for mining less than clean data. In literature \cite{he2014practical}, a hybrid model, combining \textit{GBDT} with Logistic Regression (\textit{LR}), outperforms either of these methods on their own. It treats each individual tree as a bin feature and takes the index of the leaf node an instance ends up falling in as value. Therefore, it converts a real-valued vector into a compact binary-valued vector, further included into the next linear model, i.e. \textit{LR}. Compared with \cite{he2014practical}, our proposed method employs the cross-entropy based feature representation in a deep cascaded structure, which results in strong and explainable representation ability, i.e., a traversal from root to a leaf node in next level \textit{GBDT} indicates a combination rule of certain paths on the trees from preceding level \textit{GBDT}.

\textit{gcForest}, an alternative to deep neural networks for many tasks, is proposed in literature \cite{zhou2017deep}, which employs deep forest structure to do representation learning. Specifically, it takes a multi-dimensional class vector from preceding level, together with the original feature vector, as the inputs of next level. Our proposed methods mainly have two significant differences from \textit{gcForest}:

\begin{itemize}
\item{In \textit{gcForest}, the class-specific prediction probabilities form a feature vector, further included into the next level forest after concatenating it with the original features. However, We employ \textit{GBDT} as the base unit in the proposed deep cascade structure. In addition, we use the cross-entropy in each tree of the preceding \textit{GBDT} as the feature representation for next level \textit{GBDT}. The aforementioned two points lead to a more explainable feature representation ability of the proposed method, e.g., a traversal from root to a leaf node in next level \textit{GBDT} indicates a combination rule of certain paths on the trees from preceding level \textit{GBDT}.}

\item{Compared with \textit{gcForest}, Our method takes into account mutual complementarity between strong correlation features and weak correlation features for better representation learning.}
\end{itemize}

\section{The Proposed Approach}

In this section, we firstly proposed a novel multi-Level Deep Cascade Trees (\textit{ldcTree}) to tackle the \textit{CVR} prediction problem in recommendation. Specifically, the base structure of the \textit{ldcTree} is constructed by stacking several \textit{GBDT}s sequentially, and the cross-entropy of each leaf node in the preceding \textit{GBDT} is calculated and used to form the input feature representation for the next \textit{GBDT}. Moreover, inspired by the idea of ensemble learning, an improved structure of the \textit{ldcTree} is proposed, named Ensemble \textit{ldcTree} (\textit{EldcTree}), which further encourages the model's diversity and enhances the representation ability. Finally, we notice that a small number of features contributes the majority of explanatory power while the remaining features have only a marginal contribution in \textit{GBDT} models \cite{he2014practical}, which leads to the importance of weak correlation features, especially the combination of weak correlation features, can't be revealed. Therefore, based on \textit{EldcTree}, We propose an improved Feature learning method, named \textit{F-EldcTree}. We will present the details in the following parts.

\subsection{Representation Learning based on Cascade Trees}

\begin{figure}
\centering
\includegraphics[width=0.80\linewidth]{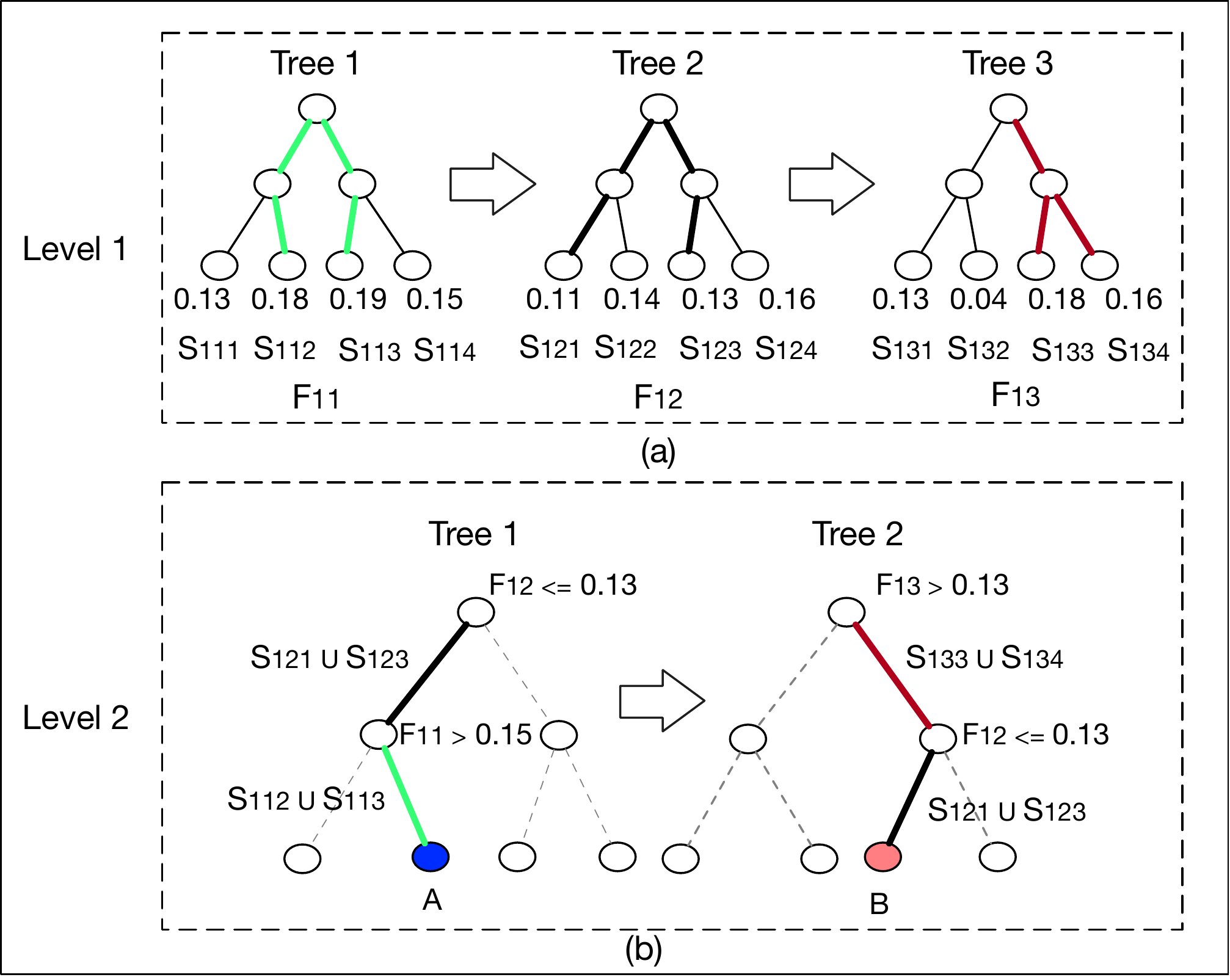}
\caption{Illustration of feature representation in the \textit{ldcTree}. (a) an exemplar \textit{GBDT} of three trees at level 1. (b) an exemplar \textit{GBDT} at the next level of two trees. Each splitting node at the level 2 corresponds a split of paths in a certain tree at level 1. Please note the corresponding colors of paths at the two levels.}
\label{fig:tree_repl2}
\end{figure}

Inspired by representation learning in deep neural network which mostly relies on the level-by-level abstraction of features, we propose a novel method named \textit{ldcTree} by employing the deep cascade tree structure. In \textit{ldcTree}, level-by-level greedily learning towards the final target is carried out. Specifically, each level with a certain number of trees included in \textit{GBDT}, receives the outputted features from its proceding level. Referring to Fig. \ref{fig:tree_repl2}, it is an illustration of \textit{ldcTree}, where it contains two levels, with each level three trees and two trees respectively. To facilitate the narration, some mathematical notations are defined as follows:

\begin{itemize}
\item{$S_{ijk}$: the cross-entropy of the $k$-th leaf node of the $j$-th tree at level $i$.}
\item{$I_{ijk}$: the number of instances falling in the $k$-th leaf node of the $j$-th tree at level $i$.}
\item{$\delta _{ijk}$: the split threshold for the k-th node of the j-th tree at the level $i$.}
\item{$L_{ij}$: the number of leaf nodes of the $j$-th tree at level $i$.}
\item{$F_{ij}$: the feature value at the j-th dimension of the feature at level $i$.}
\item{$N_{i}$: the number of trees for the \textit{GBDT} model at level $i$.}
\item{$h_{ij}(x^{n})$: the predicted probability of the n-th instance $x^{n}$ on the j-th tree at the level $i$.}
\item{$y_{n}$: the ground truth label of the n-th instance $x^{n}$: 0 or 1 in our two class \textit{CVR} problem.}
\end{itemize}
Given an instance, according to the principle of \textit{GBDT} model, each individual tree will produce a path from the root node to a leaf node. Instances are split into different paths and each leaf node will gather a certain part of them. Then, we define the cross-entropy $S_{ijk}$ at each leaf node as:

\begin{equation}
  S_{ijk}=-\frac{1}{I_{ijk}}\sum_{m=1}^{I_{ijk}}[y_{m}log(h_{ij}(x^{m}))+(1-y_{m})log(1-h_{ij}(x^{m}))]
\end{equation}
Therefore, there are $L_{ij}$ cross-entropy values on the $j$-th tree on the $i$-th level, i.e., $S_{ij1},S_{ij2},...,S_{ijL_{ij}}$ and each of them is a possible instantiation of $F_{ij}$. For all the trees in a \textit{GBDT} at the level $i$, we denote the feature representation as [$F_{i1},F_{i2},...,F_{iN_{i}}$], and use it as the inputs of the \textit{GBDT} at the level $i+1$.

Without loss of generality, assuming the feature $F_{ij}$ with the best \textit{gini} value is chosen for splitting instances on the $q$-th node of the $p$-th tree at the level $i+1$. The split threshold $\delta _{(i+1)pq}$ divides the set of $S_{ij1},S_{ij2},...,S_{ijL_{ij}}$, into two subsets $S^{+}_{ij},S^{-}_{ij}$:
\begin{equation}
S^{+}_{ij}=\bigcup_{p=1}^{L_{ij}}I(S_{ijp}>\delta _{(i+1)pq})S_{ijp}
\label{eq:s_pos}
\end{equation}

\begin{equation}
S^{-}_{ij}=\bigcup_{p=1}^{L_{ij}}I(S_{ijp}\leq \delta _{(i+1)pq})S_{ijp}
\label{eq:s_neg}
\end{equation}
where $I(z)$ is an indicator faction that outputs 1 if $z$ is true and zero otherwise.

\newtheorem{remark-MRP}{Remark}
\theoremstyle{remark}
\begin{remark-MRP}
By using cross-entropy as the basic feature representation for leaf nodes, the proposed $ldcTree$ has a clear explanation: i.e., a traversal from root to leaf nodes
in the next level \textit{GBDT} corresponds to the combination of certain traversals in the preceding \textit{GBDT}, which also leads to a distributed feature representation ability.
\end{remark-MRP}\
\emph{Explanation: }Usually, $\delta _{(i+1)pq}$ in Eq.\eqref{eq:s_pos} and Eq.\eqref{eq:s_neg} takes a value from the set of $S_{ij1},S_{ij2},...,S_{ijL_{ij}}$. The splitting process is carried out repeatedly until the stop rule for leaf nodes holds (We show an example in Fig. \ref{fig:tree_repl2}). It can be seen that, each element in $S^{+}_{ij}$ or $S^{-}_{ij}$ corresponds a path on the $j$-th tree of level $i$. Therefore, each splitting node at the level $i+1$ corresponds a split of paths in a certain tree at the preceding level $i$. Consequently, a path at the level $i+1$ corresponds a union of several paths at the preceding level $i$. It indicates the nature of the proposed $ldcTree$: \emph{a clear explanation and a distributed feature representation ability}. As shown in Fig. \ref{fig:tree_repl2}, an instance is represented as a three-dimensional feature vector $[F_{11}, F_{12}, F_{13}]$ after level 1 with three trees, where $F_{11}$ takes a value from the set of 0.13, 0.18, 0.19, 0.15, analogously for $F_{12}$ and $F_{13}$. Given $[F_{11}, F_{12}, F_{13}]$ as the inputs at level $2$, $F_{12}$ is chosen as the split feature with the split threshold $0.13$ on Tree 1, which leads to $S_{121},S_{123}\subset S^{-}_{12}$ and $S_{122},S_{124}\subset S^{+}_{12}$. And analogously, $S_{112},S_{113}\subset S^{+}_{11}$ and $S_{111},S_{114}\subset S^{-}_{11}$. Consequently, the union of paths at leaf node A can be represented as $S^{-}_{12} \cap  S^{+}_{11}$, i.e. $(S_{121}\bigcup S_{123})\bigcap (S_{112}\bigcup S_{113})$, and analogously for Tree 2.

\begin{figure}
\centering
\includegraphics[width=0.7\linewidth]{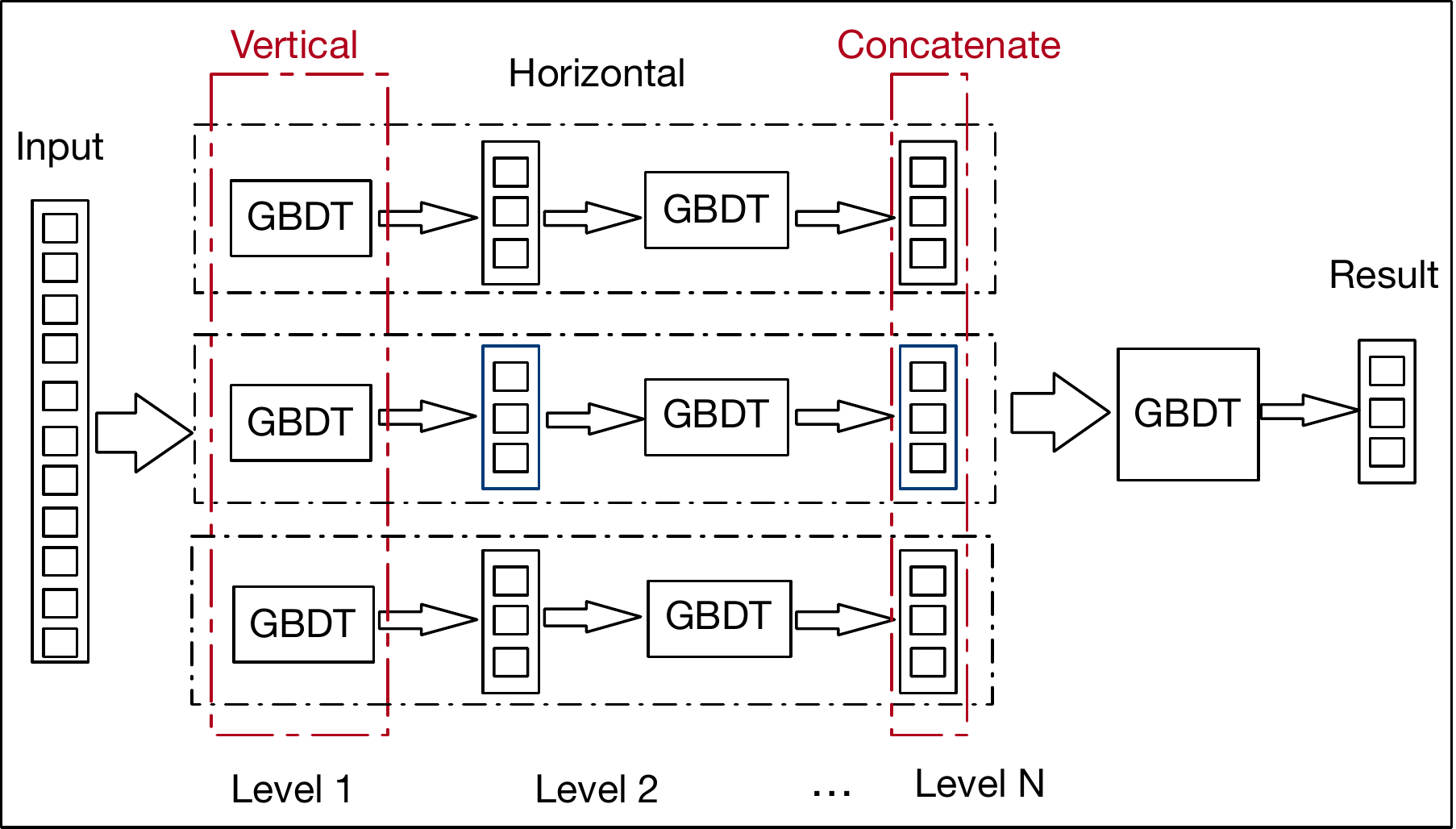}
\caption{Illustration of the structure of \textit{EldcTree}, where each horizontal part of the ensemble structure is a single \textit{ldcTree}, and each vertical part of the ensemble structure are parallel \textit{GBDT}s. The input features of the first level \textit{GBDT}s are different and randomly chosen from a feature pool, which encourages the model's diversity. The outputted features from each \textit{ldcTree} are concatenated together and used as the input features of the last \textit{GBDT} for final prediction.}
\label{fig:ldcTree}
\end{figure}

Moreover, inspired by the ensemble learning idea, we propose an Ensemble \textit{ldcTree} named \textit{EldcTree} by constructing several parallel \textit{ldcTree}s to enhance the representation ability further. As shown in Fig. \ref{fig:ldcTree}, each horizontal part of the ensemble structure is a single \textit{ldcTree}, and each vertical part of the ensemble structure are parallel \textit{GBDT}s. Their initial input features for the first level of each \textit{ldcTree} are different and chosen randomly from a raw feature pool which encourages the model's \textit{diversity}. The last level's outputted features from each \textit{ldcTree} are concatenated together and used as the input features of the last \textit{GBDT} for final prediction. Here, the \textit{diversity} in our model not only represents the nature of diverse abstracted high-level features, but also the value of inherent ensemble learning. For example, each individual \textit{Horizontal} structure randomly choose a subset of features
from the feature pool as input, which are then abstracted to different high-level features through the cascaded
structure. It inherits the idea in feature engineering by learning high-level features from different
combinations of low-level raw features. Then, the final \textit{GBDT} model, concatenating all the high-level features from preceding \textit{ldcTree} as inputs, indeed follows the ensemble learning idea for further enhancing the performance of the entire model. It is noteworthy that such an ensemble structure is naturally apt to parallel implementation and have the potential for incremental learning, e.g., the idea used in broad learning system \cite{chen2018broad}. We leave it as a future work.

\subsection{Cascade Trees Associated With Weak Correlation Features}

Though features can be chosen randomly from a given feature pool and used as the input features of each \textit{ldcTree}, features indeed have different importance for prediction. In this paper, we use the \textit{statistic Boosting Feature Importance} \cite{he2014practical}, which aims to capture the cumulative loss reduction attributable to a feature, to measure feature's importance. More specifically, a best feature is selected and split to maximize the squared error reduction during each tree node construction. Therefore, the importance of each feature is determined by summing the total reduction for itself across all the trees. Typically, a small number of features contributes the majority of explanatory power while the remaining features have only a marginal contribution in \textit{GBDT} models. Here, we regard the features contributing the majority of explanatory power as \textbf{SCF}, i.e. ``Strong Correlation Features'', while features having only a marginal contribution as \textbf{WCF}, i.e. ``Weak Correlation Features''.

\begin{figure*}
\centering
\includegraphics[width=0.7\linewidth]{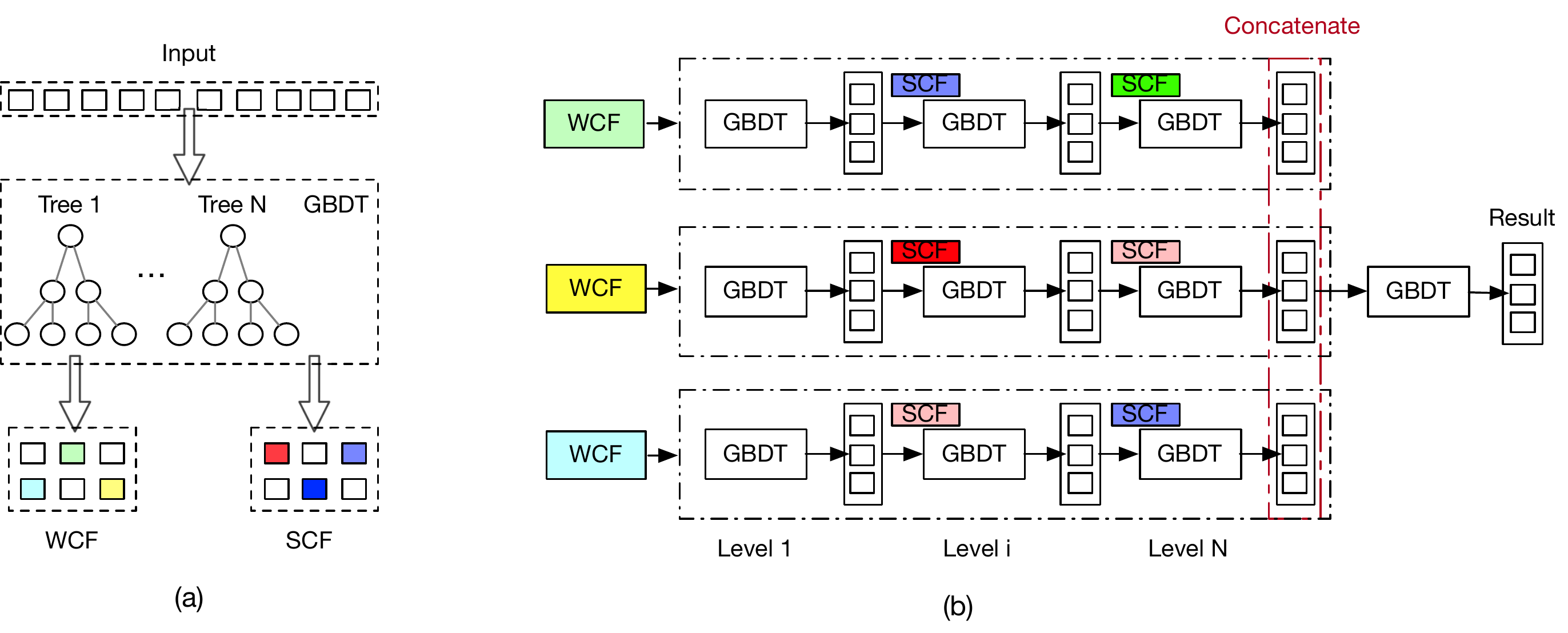}
\caption{Illustration of the proposed \textit{F-EldcTree} method, which utilizes weak correlation features and strong correlation features in a coordinated manner.}
\label{fig:weak_strong}
\end{figure*}

To take adequate use of \textit{WCF}, we proposed an improved Feature learning method based on the above \textit{EldcTree}, named \textit{F-EldcTree}, whose structure is showed in Fig. \ref{fig:weak_strong}. In Fig. \ref{fig:weak_strong}(a), a separated \textit{GBDT} is pre-trained to identify the importance of all the initial raw features, and these features are further split into two subsets, namely \textit{WCF} and \textit{SCF}. Due to our practical lessons, single feature from \textit{WCF} contributes little to final prediction results, while the combinations of these features from \textit{WCF} not. Moreover, in \textit{GBDT} models, a traversal from root to a leaf node in each tree indicates a combination rule of certain raw features. Therefore, we can resort to \textit{GBDT} models for uniting certain features from \textit{WCF} to take full advantage of them and collaborate with \textit{SCF}s to further improve the prediction accuracy.

The strategies of randomly selecting features in \textit{gcForest} \cite{zhou2017deep} convinces us its effectiveness in ensemble learning. Therefore, in Fig. \ref{fig:weak_strong}(b), features are also randomly chosen from \textit{WCF}s during the first level \textit{GBDT} training stage of each \textit{ldcTree}, which not only can learn the combination of certain \textit{WCF}s, but also encourages the model's diversity and enhances the representation ability further. For \textit{GBDT}s at the remaining levels of each \textit{ldcTree}, their input features consists of two parts: one is the representational features from the preceding level, the other is randomly chosen from the \textit{SCF}s. In this way, the \textit{F-EldcTree} starts learning features from \textit{WCF}s and combines the learned features (i.e., the combinations of \textit{WCF}s) with \textit{SCF}s little by little. Lastly, an additional \textit{GBDT} model concatenates all the representational features from \textit{ldcTree}s as the inputs for final results.

\section{Experimental results}

To evaluate the effectiveness of the proposed method, we conducted extensive experiments including off-line and online evaluations. First, we present the evaluation settings including data set preparation, evaluation metrics, a brief description of related comparison methods, and the hyper-parameters settings of the proposed method. Then, we present the experimental results of different methods on the off-line data set along with the analysis. Finally, we present experimental results of different methods for online deployment through A/B test.

\subsection{Evaluation settings}

\subsubsection{Data set preparation}
The off-line benchmark data set was constructed from the real click and purchase logs of our recommendation module in several consecutive days of December, 2017. And it consists of more than 100M instances, each of which contains user/item features and label (here, an individual id for each instance, while label is positive if purchase after clicking, or negative if no purchase after clicking). When preparing off-line benchmark data set, we divided the whole benchmark data into three disjoint parts according to the id, i.e., 40, 20 and 40 percent of the whole benchmark data for training data, validation data and test data respectively. Additionally, we extract hundreds of raw features including user features, item features and user-item cross features for each instance. For example, user features include users' ages, genders and purchasing powers, etc. Item features include items' prices, historical \textit{CVR}s and Click-through Rate (\textit{CTR}s), etc. User-item cross features include uses' historical \textit{CVR}s, preference scores, etc, on items.

\subsubsection{Evaluation metrics}
In order to comprehensively compare the performance of the propose method with other related methods, we adopt two commonly used metrics namely \textit{AUC} (Area Under Curve) and $F_1$ score based on precision and recall. Specifically, denoting all ground truth positive instances as \textit{T} and all predicted positive instances as \textit{P}, then precision and recall are defined as follows:

\begin{equation}
precision=\frac{|P\bigcap T|}{|P|}
\end{equation}

\begin{equation}
recall=\frac{|T\bigcap P|}{|T|}
\end{equation}
Then, the $F_1$ score is defined as:
\begin{equation}
F_{1}=\frac{2*precision*recall}{precision + recall}
\end{equation}

\subsubsection{Related comparison methods}
In the following experiments, we compare the proposed method with other related methods including:
\begin{itemize}

\item{\textit{Naive GBDT}: We refer to a single \textit{GBDT} model without level-by-level learning as \textit{Naive GBDT}.}

\item{\textit{GBDT + LR}: First, feature representation is trained from a \textit{GBDT} model. Then, it is used for \textit{CVR} prediction by a Logistic Regression (LR). For feature representation, it calculated a bin feature for each individual tree by taking the index of the leaf node which an instance ends up falling in as the corresponding feature value. Therefore, it converts a real-valued raw feature vector into a compact binary-valued feature vector \cite{he2014practical}.}

\item{\textit{DNN}: Referring to \cite{he2017neural}, we design a \textit{DNN} structure including three hidden layers and a prediction layer, where ReLU is used as the activation function for each hidden layer. We choose the hyper-parameters on the validation set. For instance, the number of units for each hidden layer is set as 128. Dropout rate is set as 0.5. We use the cross-entropy loss and \textit{SGD} algorithm to train this \textit{DNN} model.}

\item{\textit{gcForest}: Following our practical experience, we replace the Forests in the original \textit{gcForest} \cite{zhou2017deep} with \textit{GBDT}s. Since the \textit{CVR} prediction is a binary classification problem, a two-dimensional class-specific feature vector is obtained from each \textit{GBDT}. Then, it is used along with the raw feature vector as the inputs of next level \textit{GBDT} for learning deep feature representation further.}
\end{itemize}.

Moreover, we also evaluate the performance among our proposed \textit{ldcTree}, \textit{EldcTree} and \textit{F-EldcTree} methods for demonstrating the effectiveness of ensemble learning and the adequate use of \textit{WCF} and \textit{SCF} at corresponding levels. It should also be noted that all methods use the same raw features as inputs if not specified (e.g., here, user, item and user-item cross features are included.)

\subsubsection{Hyper-parameters settings}

\begin{table}
\centering
  \caption{Hyper-parameters in \textit{ldcTree} and \textit{EldcTree}}
  \small
  \begin{tabular}{ccl}
    \toprule
    Parameters Name&Value\\
    \midrule
    the type of loss function& logistic loss\\
    minimum instance numbers when node split & 20\\
    sampling rate of train set for each iteration & 0.6 \\
    sampling rate of features for each iteration & 0.6 \\
    the tree depth & 8\\
    the number of trees & 150\\
    learning rate& 0.01 \\
  \bottomrule
\end{tabular}
\label{tab:freq}
\end{table}

We choose the hyper-parameters of the proposed methods according to the \textit{AUC} metric on the validation set. And the main hyper-parameters of the proposed methods we used in all the following concern experiments are shown in Tab. \ref{tab:freq}. Here, we take a critical parameter ``the tree depth'' as example to illustrate the process of parameter selection in \textit{ldcTree} model. After sampling a small subset from the whole data, we train different \textit{ldcTree} models by changing the tree depth while fixing other settings. Results are shown in Tab. \ref{tab:auc_cost}. It can be seen that the time cost for each iteration increases consistently with the growth of the tree depth, and the corresponding \textit{AUC} on the validation set also increases. In addition, the time cost increases significantly when the depth increases from 8 to 10, while for the \textit{AUC} value, it grows at a snail's pace. For example, marginal gains of 0.003 are achieved by increasing the depth from 8 to 10. Therefore, to achieve a trade-off between model capacity and complexity, we set the hyper-parameter ``the tree depth'' as 8. Moreover, \textit{EldcTree} and \textit{F-EldcTree} can refer to the same hyper-parameters as \textit{ldcTree}. Finally, aforementioned naive experiments also taught us that the performance of models can't be boosted further while increasing the tree depth blindly.

\begin{table}
  \caption{The metric on Time Cost and the AUC in \textit{ldcTree}}
  \label{tab:auc_cost}
  \small
  \begin{tabular}{ccccc}
    \toprule
    Metric Name&Depth 4 & Depth 6 & Depth 8 & Depth 10\\
    \midrule
    \textit{Time Cost(s)}& 5 & 7 & 15 & 26\\
    \textit{AUC}& 0.783 & 0.789 & 0.794 & 0.797\\
  \bottomrule
\end{tabular}
\label{tab:auc_cost}
\end{table}

\subsection{Comparison results on off-line dataset}

We report the \textit{AUC} values and $F_{1}$ scores of different methods on the off-line test set. The results are shown in Tab. \ref{tab:auc} and Tab. \ref{tab:all_method}, respectively. Referring to Tab. \ref{tab:auc}, it can be seen that the \textit{GBDT + LR} method achieves a gain of 0.0039 compared to \textit{Naive GBDT} due to the additional \textit{LR} classifier. \textit{DNN} and \textit{gcForest} achieve better results than \textit{GBDT + LR}. It convinces us that strong representation features are learned due to their deep structures. Our proposed method \textit{ldcTree} achieves higher \textit{AUC} values than certain related methods(such as \textit{Naive GBDT}, \textit{GBDT+LR}, \textit{DNN}). Moreover, the proposed \textit{EldcTree} achieves better \textit{AUC} result than \textit{ldcTree} due to the idea of ensemble learning. Finally,
the further proposed \textit{F-EldcTree} achieves the best result than all the other competitive methods by taking full use of \textit{WCF} and \textit{SCF} little by little, together with the idea of ensemble learning. And the gain is nearly 0.063 compared to the initial baseline \textit{Naive GBDT}. According to our practical lessons, it should be noted that a gain of 0.01 in off-line \textit{AUC} can lead to big increment  in revenue in our online recommendation system. In conclusion, the significant gain over initial \textit{Naive GBDT} convinces the effectiveness of the proposed deep cascade structure for stronger feature representation, and the gain over \textit{gcForest} convinces the effectiveness of level-by-level learning, for example, taking the outputs of preceding level as the inputs of the next level.
Moreover, compared with \textit{EldcTree}, results of \textit{F-EldcTree} convince the idea by taking full use of weak and strong correlation features.

As for the $F_1$ score, we report several values by setting different thresholds. First, we sort all the instances in a descending order according to the predicted score. Then, we choose 3 thresholds namely top@10\%, top@20\% and top@50\% to split the predictions into positive and negative groups accordingly. Finally, we calculate the Precision, Recall and $F_{1}$ scores of these predictions at different thresholds. Results are showed in Tab. \ref{tab:all_method}. As can be seen, the proposed method achieves the best performance which is consistent with Tab. \ref{tab:auc}.

\begin{table}
\centering
  \caption{Comparison AUC results for all competitors.}
  \small
  \begin{tabular}{ccc}
    \toprule
    Method Name&AUC\\
    \midrule
    \textit{Naive GBDT}& 0.7692\\
    \textit{GBDT + LR}& 0.7731\\
    \textit{DNN}& 0.7793\\
    \textit{gcForest}& 0.7854\\
    \textit{ldcTree}& 0.7942\\
    \textit{EldcTree} & 0.8121\\
    \textit{F-EldcTree} & 0.8315\\
  \bottomrule
\end{tabular}
\label{tab:auc}
\end{table}

\begin{table*}[htbp]
\centering
\small
\caption{The Precision, Recall and $F_{1}$ score for all the competitors.}
\begin{tabular}{c|c|ccc|ccc|ccc}
\hline
\multirow{2}{*}{Method Type} &
\multirow{2}{*}{Method Name} &
\multicolumn{3}{c|}{top@10 percent} &
\multicolumn{3}{c|}{top@20 percent} &
\multicolumn{3}{c}{top@50 percent} \\
\cline{3-11}
&& Precision & Recall & F1-Score & Precision & Recall & F1-Score & Precision & Recall & F1-Score \\
\hline
\multirow{4}*{Compare Methods}
&\textit{Naive GBDT} &5.75\%&36.39\%&9.93\%&4.16\%&51.69\%&7.70\%&2.42\%&76.57\%&4.68\%\\
\cline{2-11}
&\textit{GBDT + LR} & 6.39\%&37.40\%&10.92\%&4.63\%&54.16\%&8.52\%&2.68\%&78.71\%&5.19\%\\
\cline{2-11}
&\textit{DNN} & 6.79\%&38.32\%&11.54\%&4.92\%&55.51\%&9.03\%&2.85\%&80.96\%&5.51\%\\
\cline{2-11}
&\textit{gcForest} & 6.82\%&39.03\%&11.61\%&4.94\%&56.52\%&9.08\%&2.87\%&82.14\%&5.54\%\\
\hline
\multirow{4}*{Our Methods}
&\textit{ldcTree} & 6.85\%&39.86\%&11.69\%&4.96\%&57.72\%&9.13\%&2.88\%&82.92\%&5.56\%\\
\cline{2-11}
&\textit{EldcTree} & 7.44\%&41.49\%&12.62\%&5.39\%&61.81\%&9.91\%&3.13\%&87.31\%&6.04\%\\
\cline{2-11}
&\textit{F-EldcTree} & \textbf{8.16\%}&\textbf{43.91\%}&\textbf{13.76\%}&\textbf{5.91\%}&\textbf{63.58\%}&\textbf{10.81\%}&\textbf{3.43\%}&\textbf{92.40\%}&\textbf{6.61\%}\\
\hline
\end{tabular}
\label{tab:all_method}
\end{table*}

\subsection{Online evaluation results}

\begin{table}
\centering
  \caption{The Effectiveness of level-by-level learning.}
  \small
  \begin{tabular}{ccccc}
    \toprule
     Model Name&Day 1&Day 2&Day 3&Day 4\\
    \midrule
    \textit{Naive GBDT}& 100\%& 100\%& 100\%& 100\%\\
    \textit{ldcTree}& \textbf{104.3\%}& \textbf{104.1\%}& \textbf{103.8\%}& \textbf{103.9\%}\\
    \textit{EldcTree}& \textbf{107.1\%}& \textbf{107.4\%}& \textbf{106.3\%}& \textbf{106.8\%}\\
  \bottomrule
\end{tabular}
\label{tab:level_level}
\end{table}

Next, we firstly present the effectiveness of level-by-level learning through online contrastive experiments. Then, we further demonstrate the effectiveness of \textit{F-EldcTree} by taking full advantage of \textit{WCF} and \textit{SCF} compared with other competitors. It should also be noted that all comparison methods use the same input features if not specified (e.g., here, user, item and user-item cross features are included.). In addition, we fix all the other online recommendation modules unchanged except the \textit{CVR} module.

\subsubsection{The Effectiveness of level-by-level Learning}

For demonstrating the effectiveness of the proposed level-by-level learning, we implement \textit{ldcTree} and \textit{EldcTree} methods with the same features from \textit{Naive GBDT}. After deploying them in the online recommendation system, we record four days' purchase logs and calculate the relative increasement in \textit{CVR}. The A/B test results are showed in Tab. \ref{tab:level_level}. It can be seen the proposed \textit{ldcTree} method achieves more than 4\% gain of \textit{CVR} averagely, while \textit{EldcTree} more than 7\%. In addition, after analyzing the difference between the structures of the two methods, we find that the gain mainly comes from the stronger feature representation ability of the proposed deep cascade structure in \textit{EldcTree}. It is consistent with the experimental results on the aforementioned off-line data set.

\begin{figure}
\centering
\includegraphics[width=0.8\linewidth]{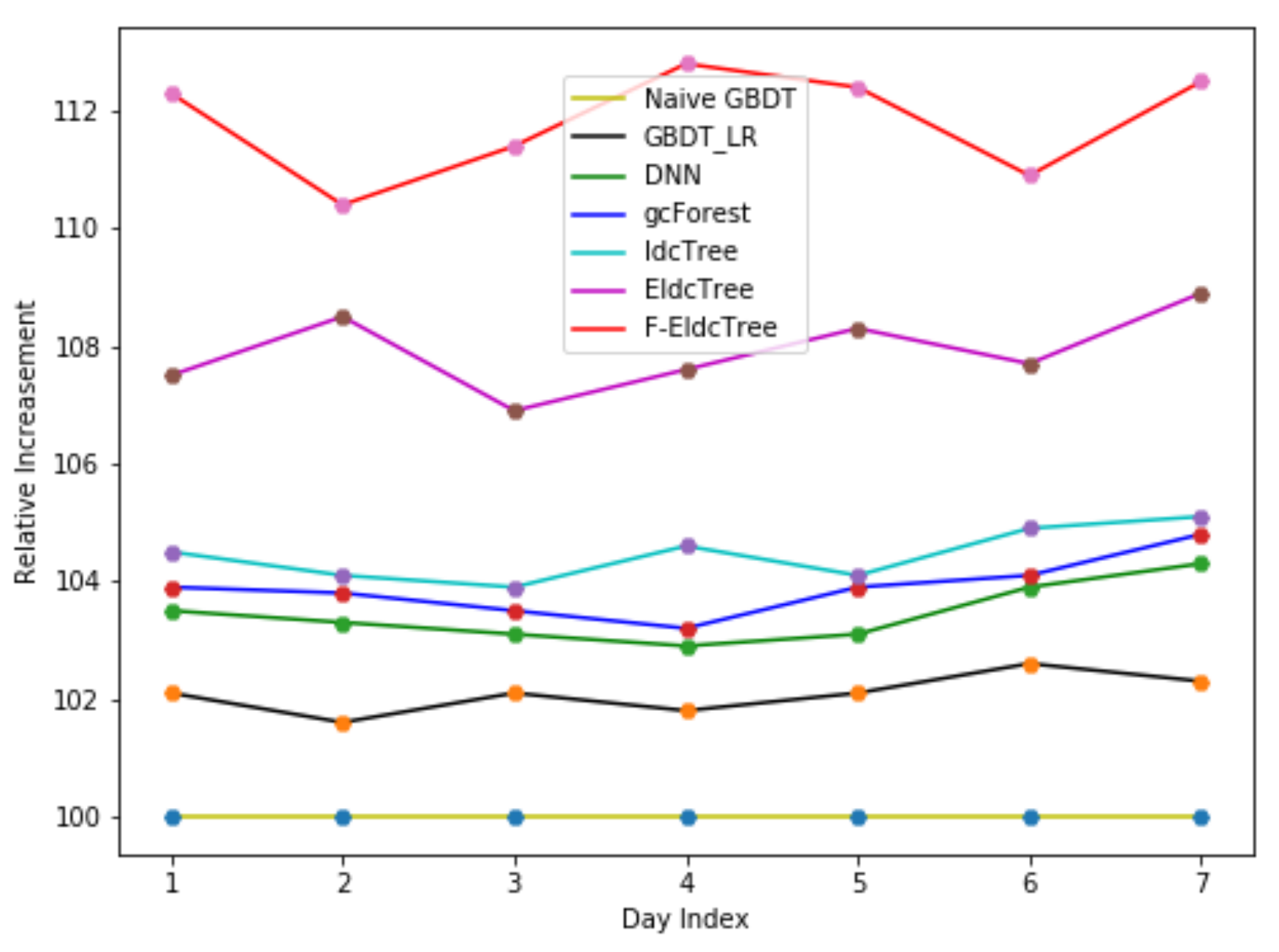}
\caption{The online A/B test results on \textit{CVR}. The initial baseline model marked in yellow is based on \textit{Naive GBDT}. Here, all the comparison methods resort to the same features including user, item, and user-item cross features.}
\label{fig:online_result}
\end{figure}

\subsubsection{The Effectiveness of \textit{F-EldcTree}.}

After demonstrating the effectiveness of utilizing level-by-level learning through online A/B experiments, we employed \textit{F-EldcTree} for online environment. In addition, the features for other competitive methods are exactly the same with \textit{F-EldcTree}. The A/B test results are showed in Fig. \ref{fig:online_result}, where \textit{gcForest} and \textit{DNN} achieve better results than \textit{Naive GBDT} due to their deep feature representation abilities. As for the proposed method, it achieves the best result, i.e. 12 percent increment in \textit{CVR} among all the methods.

In a nutshell, considering the experimental results from both off-line and online tests, we conclude that the proposed method has a stronger feature representation ability due to its deep cascade structure and the adequate use of \textit{WCF}s and \textit{SCF}s at corresponding levels. Moreover, these two distinct characteristics enables the learned features to have a clear explanation as depicted in Section 3.1.

\section{Conclusions and future work}

In this paper, we introduce effective and efficient distributed feature learning methods \textit{ldcTree} and its extension \textit{EldcTree}, which have a deep cascade structure by stacking several \textit{GBDT} units sequentially. By using a cross-entropy based feature representation, it leads to a clear explanation and a distributed feature representation ability. Moreover, after taking into account mutual complementarity between strong correlation features and weak correlation features under the ensemble learning framework, the proposed method \textit{F-EldcTree} achieves the best performance in both off-line and on-line experiments. Specifically, we successfully deploy the proposed method online in our E-commerce platform, it achieves a significant improvement compared with the previously baseline, i.e. 12 percent increment in \textit{CVR}. Our methods have small training cost and are naturally apt to parallel implementation. In addition, it is promising to be applicable for other online advertising scenarios.

Future work may include the following two directions: 1) Incorporating more features, such as information from the parent nodes and sibling nodes for learning stronger feature representation. 2) Studying the end-to-end training method for jointly feature learning and classifying based on the proposed deep cascade tree structure.

\section*{Acknowledgment}
This work was partly supported by the National Natural Science Foundation of China (NSFC) under Grants 61806062 and 61751304.


\bibliographystyle{aaai}

\end{document}